\def\year{2017}\relax
\documentclass[letterpaper]{article}

\usepackage{aaai17}
\usepackage{times}
\usepackage{helvet}
\usepackage{courier}
\usepackage{amsfonts}       
\usepackage{microtype}      

\usepackage{amsmath}
\usepackage{tabularx}
\usepackage[pdftex]{graphicx}

\begin{document}
\title{Variational Autoencoder for Semi-supervised Text Classification}

%

\author{ Weidi Xu \and Haoze Sun \and Chao Deng \and Ying Tan \\
  Key Laboratory of Machine Perception (Ministry of Education), \\
  School of Electronics Engineering and Computer Science, Peking University, Beijing, 100871, China \\
  wead\_hsu@pku.edu.cn, pkucissun@foxmail.com, cdspace678@pku.edu.cn, ytan@pku.edu.cn \\
}

\maketitle
\begin{abstract}
Although semi-supervised variational autoencoder (\emph{SemiVAE}) works in image classification task, it fails in text classification task if using vanilla LSTM as its decoder.
From a perspective of reinforcement learning, it is verified that the decoder's capability to distinguish between different categorical labels is essential.
Therefore, \emph{Semi-supervised Sequential Variational Autoencoder} (\emph{SSVAE}) is proposed, which increases the capability by feeding label into its decoder RNN at each time-step.
Two specific decoder structures are investigated and both of them are verified to be effective.
Besides, in order to reduce the computational complexity in training, a novel optimization method is proposed, which estimates the gradient of the unlabeled objective function by sampling, along with two variance reduction techniques.
Experimental results on Large Movie Review Dataset (IMDB) and AG's News corpus show that the proposed approach significantly improves the classification accuracy compared with pure-supervised classifiers, and achieves competitive performance against previous advanced methods.
State-of-the-art results can be obtained by integrating other pretraining-based methods.

\end{abstract}

\section{Introduction}

\noindent Semi-supervised learning is a critical problem in the text classification task due to the fact that the data size nowadays is increasing much faster than before, while only a limited subset of data samples has their corresponding labels.
Therefore lots of attention has been drawn from researchers over machine learning and deep learning communities, giving rise to many semi-supervised learning methods~\cite{socher2013recursive,dai2015semi}.

Variational autoencoder is recently proposed by \citeauthor{kingma2013auto,rezende2014stochastic}, and it has been applied for semi-supervised learning~\cite{kingma2014semi,maaloe2016auxiliary}, to which we refer as \emph{SemiVAE}.
Although it has shown strong performance on image classification task, its application in sequential text classification problem has been out of sight for a long time.
Since variational autoencoder has been verified to be effective at extracting global features from sequences (e.g., sentiment, topic and style)~\cite{bowman2015generating}, it is also promising in the semi-supervised text classification task.

In this paper, \emph{Semi-supervised Sequential Variational Autoencoder} (\emph{SSVAE}) is proposed for semi-supervised sequential text classification.
The \emph{SSVAE} consists of a Seq2Seq structure and a sequential classifier. In the Seq2Seq structure, the input sequence is firstly encoded by a recurrent neural network, e.g., LSTM network~\cite{hochreiter1997long} and then decoded by another recurrent neural network conditioned on both latent variable and categorical label.
However, if the vanilla LSTM network is adopted as the decoder, the \emph{SSVAE} will fail to make use of unlabeled data and result in a poor performance.

The explanation is given by carefully analyzing the gradient of the classifier from a perspective of reinforcement learning (RL), which reveals how the classifier is driven by the decoder using unlabeled data.
By comparing the gradient of the classifier w.r.t. unlabeled objective function to REINFORCE algorithm~\cite{williams1992simple}, we realize that only if the decoder is able to make difference between correct and incorrect categorical labels, can the classifier be reinforced to improve the performance.
Vanilla LSTM setting will mislead the decoder to ignore the label input and hence fails in the sequence classification task.

To remedy this problem, the influence of categorical information is increased by feeding label to the decoder RNN at each time step.
This minor modification turns out to bring \emph{SSVAE} into effect.
Specifically, we made an investigation on two potential conditional LSTM structures.
Experimental results on IMDB and AG's News corpus show that their performances are close and both of them are able to outperform pure-supervised learning methods by a large margin.
When using only 2.5K labeled IMDB samples, 10.3\% classification error can still be obtained, which outperforms supervised LSTM by 7.7\%.
The better one is able to achieve very competitive results compared with previous advanced methods.
Combined with pretraining method, our model can obtain the current best results on IMDB dataset.
Although LSTM is utilized as the classifier in this paper, it should be noted that the classifier can be easily replaced by other more powerful models to achieve better results.

In addition, motivated by the aforementioned interpretation, we reduce the computational complexity of \emph{SemiVAE} by estimating the gradient of unlabeled objective function using sampling.
In order to reduce the high variance caused by sampling, the baseline method from the RL literature is adopted.
For \emph{SSVAE}, two kinds of baseline methods are studied, with which the training becomes stable.

In summary our main contributions are:
\begin{itemize}
	\item We make the \emph{SSVAE} effective by using conditional LSTM that receives labels at each step,
    and give the explanation from the RL perspective.
	Two plausible conditional LSTMs are investigated.
	\item We propose an optimization method to reduce the computational complexity of \emph{SSVAE} via sampling.
	And two different baseline methods are proposed to reduce the optimization variance.
	By sampling with these baselines, the model can be trained faster without loss of accuracy.
    \item We demonstrate the performance of our approach by providing competitive results on IMDB dataset and AG's news corpus. 
	Our model is able to achieve very strong performance against current models.
\end{itemize}

The article is organized as follows. In the next section, we introduce several related works. And then our model is presented in section~\ref{sec:model}. In section~\ref{sec:exp}, we obtain both quantitative results and qualitative analysis of our models. At last we conclude our paper with a discussion.

\section{Preliminaries}
\subsection{Semi-supervised Variational Inference}
\label{sec:semivae}
\citeauthor{kingma2014semi} firstly introduced a semi-supervised learning method based on variational inference.
The method consists of two objectives for labeled and unlabeled data.
Given a labeled data pair $(x, y)$, the evidence lower bound with corresponding latent variable $z$ is:
\begin{equation}
\begin{split}
\log p_{\theta}(x,y) & \ge \mathbb{E}_{q_{\phi}(z|x,y)}[\log p_{\theta}(x|y, z)] + \log p_\theta(y) \\
			    & - D_{KL}(q_{\phi}(z|x,y) || p(z))  = -\mathcal{L}(x,y) \,,
\end{split}
\label{equ:lab}
\end{equation}
where the first term  is the expectation of the conditional log-likelihood on latent variable $z$, and the last term is the Kullback-Leibler divergence between the prior distribution $p(z)$ and the learned latent posterior $q_{\phi}(z|x,y)$.

For the unlabeled data, the unobserved label $y$ is predicted from the inference model with a learnable classifier $q_{\phi}(y|x)$. The lower bound is hence:
\begin{equation}
\begin{split}
\log p_{\theta}(x) & \ge \sum_{y}q_{\phi}(y|x)(-\mathcal{L}(x,y))+ \mathcal{H}(q_{\phi}(y|x)) \\
			 & = -\mathcal{U}(x)\,.
\end{split}
\label{equ:unl}
\end{equation}

The objective for entire dataset is now:
\begin{equation}
\begin{split}
J & = \sum_{(x,y) \in S_{l}}\mathcal{L}(x,y) + \sum_{x \in S_{u}}\mathcal{U}(x) \\
 & + \alpha \mathbb{E}_{(x,y) \in S_{l}}[-\log q_{\phi}(y|x)] \,,
\end{split}
\label{equ:com}
\end{equation}
where $S_l$ and $S_u$ are labeled and unlabeled data set respectively, $\alpha$ is a hyper-parameter of additional classification loss of labeled data.

\subsection{Semi-supervised Variational Autoencoder}
This semi-supervised learning method can be implemented by variational autoencoder~\cite{kingma2013auto,rezende2014stochastic} (\emph{SemiVAE}).
The \emph{SemiVAE} is typically composed of three main components: an encoder network, a decoder network and a classifier, corresponding to $q_\phi(z|x,y)$, $p_\theta(x|y,z)$ and $q_\phi(y|x)$.

In the encoder network, each data pair $(x,y)$ is encoded into a soft ellipsoidal region in the latent space, rather than a single point, i.e., the distribution of $z$ is parameterized by a diagonal Gaussian distribution $q_{\phi}(z|x,y)$:
\begin{align}
\hat{x} &= f_{enc}(x) \,,\\
q_\phi(z|x,y) &= \mathcal{N}(\mu(\hat{x},y), diag(\sigma^2(\hat{x}, y))) \,,\\
z &\sim q_\phi(z|x,y) \,.
\end{align}

The decoder is a conditional generative model that estimates the probability of generating $x$ given latent variable $z$ and categorical label $y$:
\begin{align}
p_\theta(x|y,z) &= D(x| f_{dec}(y, z)) \,,
\end{align}
where $f_{dec}(y, z)$ is used to parameterize a distribution $D$, typically a Bernoulli or Gaussian distribution for image data.

In the applications, $f_{enc}(\cdot)$, $f_{dec}(\cdot)$ and the classifier $q_{\phi}(y|x)$ can be implemented by various models, e.g., MLP or CNN networks~\cite{maaloe2016auxiliary,yan2015attribute2image}.
Overall, the \emph{SemiVAE} is trained end-to-end with reparameterization trick \cite{kingma2013auto,rezende2014stochastic}.

\section{Sequential Variational Autoencoder for Semi-supervised Learning}
\label{sec:model}
\begin{figure}
\centering
\includegraphics[width=3.25in]{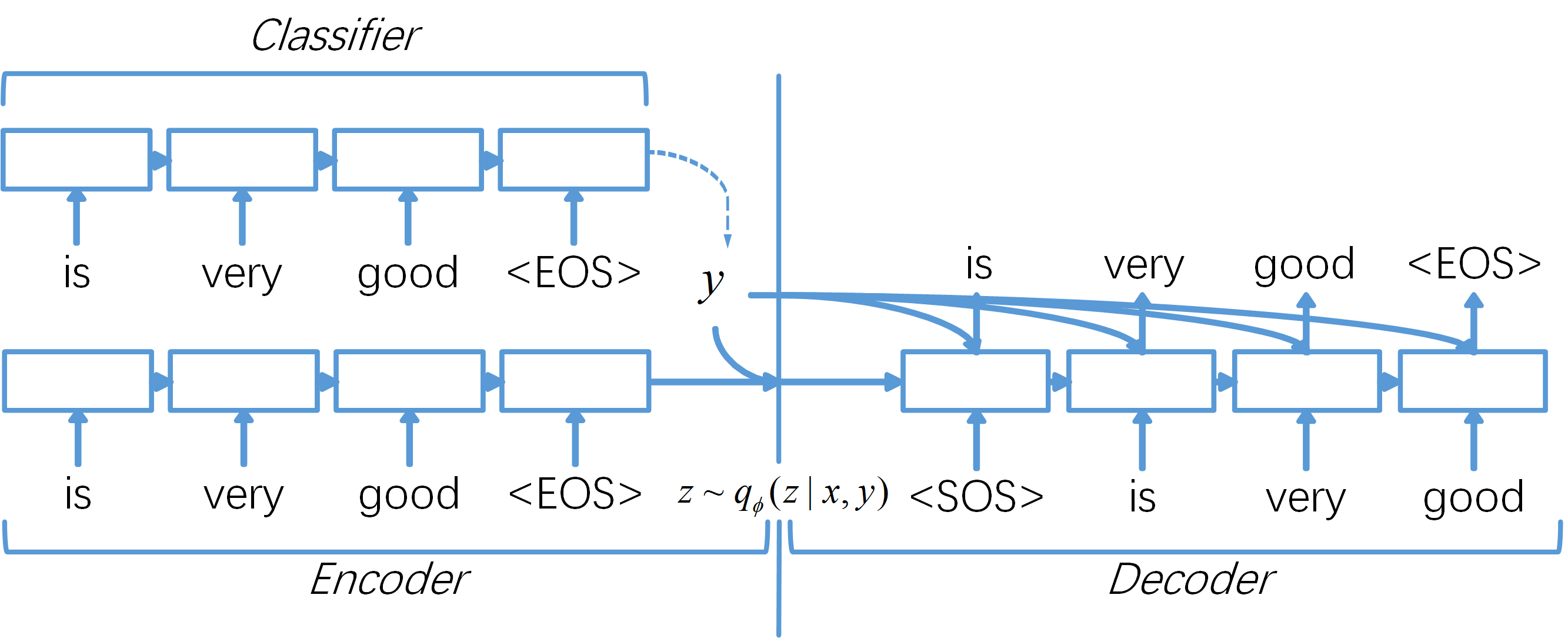}
\caption{This is the sketch of our model.
{\bf Left Bottom}: The sequence is encoded by a recurrent neural network.
The encoding and the label $y$ are used to parameterize the the posterior $q_\phi(z|x,y)$.
{\bf Right}: A sample $z$ from the posterior $q_\phi(z|x, y)$ and label $y$ are passed to the generative network which estimates the probability $p_\theta(x|y,z)$.
{\bf Left Top}: When using unlabeled data, the distribution of $y$ is provided by the sequential classifier (dashed line).
}
\label{fig:semi-sent}
\end{figure}

Based on \emph{SemiVAE}, we propose \emph{Semi-supervised Sequential Variational Autoencoder} (\emph{SSVAE}) for semi-supervised sequential text classification, sketched in Fig.~\ref{fig:semi-sent}.
In contrast to the implementation for image data, the sequential data is instead modelled by recurrent networks in our model.
Concretely, the encoder $f_{enc}(\cdot)$ and the classifier $q_\phi(y|x)$ are replaced by LSTM networks.

\subsection{Learning from Unlabeled Data}
However, problem occurs if the vanilla LSTM is used for $f_{dec}(\cdot)$ in the decoder, i.e., the latent variable $z$ and the categorical label $y$ are concatenated as the initial state for a standard LSTM network and the words in $x$ are predicted sequentially.
With this setting, the resulting performance is poor and the training is very unstable (cf. Sec.~\ref{sec:exp}).

To obtain a theoretical explanation, the gradient of the classifier $q_\phi(y|x; w_c)$, parameterized by $w_c$, is investigated.
Its gradient w.r.t. Equ.~\ref{equ:com} consists of three terms:
\begin{equation}
\begin{split}
&\Delta  w_c =  \sum_{(x,y) \in S_l} \alpha \nabla_{w_c}\log q_\phi(y|x;{w_c})  \\
&+  \sum_{x \in S_u} \nabla_{w_c} \mathcal{H}(q_{\phi}(y|x;{w_c})) \\
&  + \sum_{x \in S_u}  \mathbb{E}_{ q_\phi(y|x;{w_c})} [(-\mathcal{L}(x,y)) \nabla_{w_c} \log q_\phi(y|x;{w_c})] \,.
\end{split}
\end{equation}
The first term comes from the additional classification loss of labeled data, and the other two terms come from the unlabeled objective function (Equ.~\ref{equ:unl}).
The first term is reliable as the gradient is provided by a standard classifier $q_\phi(y|x)$ with labeled data.
The second term is a regularization term and it is negligible.
The third term, with the summation omitted,
\begin{equation}
\mathbb{E}_{ q_\phi(y|x;{w_c})} [-\mathcal{L}(x,y) \nabla_{w_c} \log q_\phi(y|x;{w_c})] \,,
\label{equ:exp}
\end{equation}
is the expectation of $\nabla_{w_c} \log q_\phi(y|x;{w_c})$ on classifier's prediction, multiplied by $-\mathcal{L}(x,y)$.
Since the evidence lower bound $-\mathcal{L}(x,y)$ largely determines the magnitude of gradient for each label $y$, it is supposed to play an important role in utilizing  the information of unlabeled data.

To verify this assumption, we investigate the term (Equ.~\ref{equ:exp}) by analogy to REINFORCE algorithm. It adapts the parameters of a stochastic model to maximize the external reward signal which depends on the model's output.
Given a policy network $P(a|s;\lambda)$, which gives the probability distribution of action $a$ in current state $s$, and a reward signal $r(s,a)$, REINFORCE updates the model parameters using the rule:
\begin{equation}
\Delta \lambda \propto \mathbb{E}_{P(a|s;\lambda)}[r(s, a) \nabla _ \lambda \log P(a|s;\lambda)],
\label{equ:reinforce}
\end{equation}
which has the same format with Equ.~\ref{equ:exp}.
Comparing Equ.~\ref{equ:exp} to Equ.~\ref{equ:reinforce}, the classifier $q_\phi(y|x)$ can be seen as the policy network while the variational autoencoder gives the reward signal $-\mathcal{L}(x,y)$.
Actually, the \emph{SemiVAE} can be seen as a generative model with continuous latent variable $z$ and discrete latent variable $y$, combining both variational autoencoder~\cite{kingma2013auto} and neural variational inference learning (NVIL)~\cite{mnih2014neural}.
The whole model is guided by labeled data and reinforced using unlabeled data.

A prerequisite for RL is that the rewards between actions should make difference, i.e., more reward should be given when the agent takes the right action.
Similarly, in the \emph{SemiVAE}, only if $-\mathcal{L}(\cdot,y)$ can distinguish between correct and incorrect labels, can the classifier be trained to make better predictions.
And since $-\mathcal{L}(x,y)$ is largely determined by the conditional generative probability $p_\theta(x|y,z)$, it requires us to design a conditional generative model that has to be aware of the existence of label $y$.

In vanilla LSTM setting, the label is fed only at the first time step. And it is found that the model tends to ignore the class feature $y$, because minimizing the conditional likelihood of each class according to language model (i.e., predicting next word according to a small context window) is the best strategy to optimize the objective function (Equ.~\ref{equ:unl}).

\subsection{Conditional LSTM Structures}
To remedy this problem, the influence of label is increased by proposing a slight modification to the decoder RNN, i.e., feeding label $y$ at each time step as in \cite{wen2015semantically,ghosh2016contextual}.
Although this kind of implementation is simple, it turns out to bring the \emph{SSVAE} into effect.

This paper studies two potential conditional LSTM structures.
The first one concatenates word embedding and label vector at each time-step, which is widely used in \cite{ghosh2016contextual,serban2016building}. We call this structure CLSTM-I and its corresponding model \emph{SSVAE-I}.

The second conditional LSTM network is motivated by \citeauthor{wen2015semantically}. It is defined by the following equations:

\begin{align}
i_t &= \sigma(W_{wi}w_t+W_{hi}h_{t-1})\,, \\
f_t &= \sigma(W_{wf}w_t+W_{hf}h_{t-1})\,, \\
o_t &= \sigma(W_{wo}w_t+W_{ho}h_{t-1})\,, \\
\hat{c}_t &= \tanh(W_{wc}w_t+W_{hi}h_{t-1})\,, \\
c_t &= f_t\otimes c_{t-1}+i_t \otimes \hat{c}_t+ \tanh (W_{yc}y)\,, \label{equ:sclstm}\\
h_t &= o_t\otimes \tanh(c_t)\,,
\end{align}
where the equations are the same as in standard LSTM networks except that Equ.~\ref{equ:sclstm} has an extra term about $y$.
The label information is directly passed to the memory cell, without the process of four gates in LSTM.
This structure is denoted as CLSTM-II and the model with this structure is called \emph{SSVAE-II}.

\subsection{Optimizing via Sampling}
\label{sec:sample}
A limitation of the \emph{SemiVAE} is that they scale linearly in the number of classes in the data sets~\cite{kingma2014semi}.
It is an expensive operation to re-evaluate the generative likelihood for each class during training.
Actually, the expectation term in Equ.~\ref{equ:exp} can be estimated using Monte Carlo sampling from the classifier to reduce the computational complexity of \emph{SemiVAE}.

However, the variance of the sampling-based gradient estimator can be very high due to the scaling of the gradient inside the expectation by a potentially large term.
To reduce the variance, the baseline method~\cite{williams1992simple,weaver2001optimal}, which has been proven to be very efficient for reinforcement learning tasks, is adopted.
The baseline can be added without changing the expect gradient.
With baseline $b$ introduced, the Equ.~\ref{equ:exp} is transformed to:
\begin{equation}
\frac{1}{K}\sum^K_{k=1}[(-\mathcal{L}(x,y^{(k)}) - b(x)) \nabla_{w_c} \log q_\phi(y^{(k)}|x;{w_c})]\,,
\label{equ:sample_b}
\end{equation}
where $y^{(k)}\sim q_\phi(y|x;{w_c})$.
We investigate two kinds of baseline methods in this paper for \emph{SSVAE}:
\begin{itemize}
	\item \textbf{S1}
	Since the term $\log p_\theta(x|y,z)$ in $-\mathcal{L}(x,y)$ is approximately proportional to the sentence length, we implement a sequence-length-dependent baseline $b(x) = c|x|$, where $|x|$ stands for the length of input sequence. During the training, the scalar $c$ is learned by minimize MSE $(\log p_\theta(x|y,z)/{|x|} - c)^{2}$. In practice, the $\log p_\theta(x|y,z)$ is divided by $|x|$ and use $c$ directly as the baseline to reduce the variance introduced by various sentence lengths.
	\item \textbf{S2}
    The second one samples $K \geq 2$ labels and simply use the averaged $-\mathcal{L}(x,\cdot)$ as the baseline, i.e., $ b(x) = \frac{1}{K}\sum^K_{k=1} -\mathcal{L}(x,y^{(k)})$.
	Although it is a little more computationally expensive, this baseline is more robust.
\end{itemize}
To avoid confusion, the S1 and S2 tags are used to indicate that the model is trained by sampling using two different baselines, while the \emph{SSVAE-I,II} are two implementations of \emph{SSVAE} using two different conditional LSTMs.

\section{Experimental Results and Analysis}
\label{sec:exp}

The system was implemented using Theano~\cite{Bastien-Theano-2012,bergstra+al:2010-scipy} and Lasagne~\cite{dieleman2015lasagne}.
And the models were trained end-to-end using the ADAM~\cite{kingma2014adam} optimizer with learning rate of 4e-3.
The cost annealing trick \cite{bowman2015generating,kaae2016train} was adopted to smooth the training by gradually increasing the weight of KL cost from zero to one.
Word dropout~\cite{bowman2015generating} technique is also utilized and the rate was scaled from 0.25 to 0.5 in our experiments.
Hyper-parameter $\alpha$ was scaled from 1 to 2.
We apply both dropout~\cite{srivastava2014dropout} and batch normalization~\cite{ioffe2015batch} to the output of the word embedding projection layer and to the feature vectors that serve as the inputs and outputs to the MLP that precedes the final layer.
The classifier was simply modelled by a LSTM network.
In all the experiments, we used 512 units for memory cells, 300 units for the input embedding projection layer and 50 units for latent variable $z$.

\subsection{Benchmark Classification}

This section will show experimental results on Large Movie Review Dataset (IMDB)~\cite{maas-EtAl:2011:ACL-HLT2011} and AG's News corpus~\cite{zhang2015character}.
The statistic of these two datasets is listed in Table~\ref{tab:sta}.
The data set for semi-supervised learning is created by shifting labeled data into unlabeled set.
We ensure that all classes are balanced when doing this, i.e., each class has the same number of labeled points.
In both datasets we split 20\% samples from train set as valid set.

\begin{table}[htp]
\small
\center
\caption{The statistic for IMDB and AG's News dataset}
\begin{tabular}{c c c c c c}
\hline
\textbf{Dataset} & \textbf{\#labeled} & \textbf{\#unlabeled} & \textbf{\#testset}  & \textbf{\#classes}\\
\hline
IMDB & 25K & 50K & 25K & 2 \\
AG's News & 120K & 0 & 7.6k &4 \\
\hline
\end{tabular}
\label{tab:sta}
\end{table}

Table~\ref{tab:imdb_detail} and Table~\ref{tab:ag_detail} show classification results on IMDB and AG's News datasets respectively.
The model using vanilla LSTM, referred as \emph{SSVAE-vanilla}, fails to improve the classification performance.
In contrast, our models, i.e., \emph{SSVAE-I} and \emph{SSVAE-II}, are able to outperform pure-supervised LSTM by a large margin, which verifies the \emph{SSVAE} as a valid semi-supervised learning method for sequential data.
With fewer labeled samples, more improvement can be obtained.
When using 2.5K labeled IMDB samples, 10.3\% classification error can still be obtained, in contrast to 10.9\% error rate using full 20K labeled data for supervised LSTM classifier.

In addition we compare our models with previous state-of-the-art pretraining-based method~\cite{dai2015semi}.
Since their codes have not been published yet, the LM-LSTM and SA-LSTM models were re-implemented.
Although the LM-LSTM was successfully reproduced and equivalent performance reported in their paper was achieved, we are unable to reproduce their best results of the SA-LSTM.
Therefore, the LM-LSTM was used as a baseline for this comparison.
Experimental results show the \emph{SSVAE}s perform worse than LM-LSTM, indicating that pretraining is very helpful in practice, considering the difficulty in optimizing the recurrent networks.
Fortunately, since the classifier is separated in \emph{SSVAE}, our method is compatible with pretraining methods.
When integrating LM-LSTM, additional improvement can be achieved and the model obtains a tie result to the state-of-the-art result.
A summary of previous results on IMDB dataset are listed in Table~\ref{tab:imdb}, including both supervised and semi-supervised learning methods.
It should be noted that the classifier in our model can be easily replaced with other more powerful methods to get better results.
Since only a subset of AG's News corpus is used as labeled data, there is no other comparative results on AG's News corpus.

\begin{table}[ht]
\center
\small
\caption{Performance of the methods with different amount of labeled data on IMDB dataset. LM denotes that the classifier is initialized by LM-LSTM.}
\begin{tabular}{l r r r r}
\hline
\textbf{Method} & \textbf{2.5K} & \textbf{5K} & \textbf{10K} &  \textbf{20K}\\
\hline
 LSTM    & 17.97\% & 15.67\% & 12.99\% &  10.90\% \\
 SSVAE-vanilla &17.76\% & 15.81\% & 12.54\% & 11.86\%\\
 SSVAE-I   & 10.38\%  & 9.93\% & 9.61\% & 9.37\% \\
 SSVAE-II & \textbf{10.28\%} & \textbf{9.50\%} & \textbf{9.40\%}  &  \textbf{8.72\%} \\
\hline
 LM-LSTM & 9.41\% & 8.90\% & 8.45\% &  7.65\% \\
 SSVAE-II,LM & \textbf{8.61\%} & \textbf{8.24\%} & \textbf{7.98\%} &  \textbf{7.23\%} \\
\hline
SSVAE-II,S1 & 16.87\% & 15.28\% & 11.62\% & 9.75\% \\
SSVAE-II,S1,LM & \textbf{9.40\%} & \textbf{9.00\%} & \textbf{8.00\%} & \textbf{7.60\%} \\
\hline
\end{tabular}
\label{tab:imdb_detail}
\end{table}

\begin{table}[ht]
\small
\center
\caption{Performance of the methods with different amount of labeled data on AG's News dataset.}
\begin{tabular}{l r r r}
\hline
\textbf{Method} & \textbf{8K} & \textbf{16K} & \textbf{32K} \\
\hline
 LSTM    & 12.74\% & 10.97\% & 9.28\% \\
 SSVAE-vanilla & 12.69\% & 10.62\% & 9.49\% \\
 SSVAE-I  & 10.22\%  & 9.32\% & 8.54\%\\
 SSVAE-II & \textbf{9.71\%} & \textbf{9.12\%} & \textbf{8.30\%}\\
\hline
 LM-LSTM  & 9.37\% & 8.51\% & 7.99\%\\
 SSVAE-II,LM & \textbf{8.97\%} & \textbf{8.33\%} & \textbf{7.60\%}\\
\hline
SSVAE-II,S1 & 11.92\% & 10.59\% & 9.28\% \\
SSVAE-II,S2 & 9.89\% & 9.25\% & 8.49\% \\
SSVAE-II,S1,LM & 9.74\% & 8.92\% & 8.00\% \\
SSVAE-II,S2,LM & \textbf{9.05\%} & \textbf{8.35\%} & \textbf{7.68\%} \\
\hline
\end{tabular}
\label{tab:ag_detail}
\end{table}

\begin{table}
\small
\center
\caption{Performance of the methods on the IMDB sentiment classification task.}
\begin{tabular}{l r}
\hline
\textbf{Model} & \textbf{Test error rate}\\
\hline
LSTM \shortcite{dai2015semi}               & 13.50\%       \\
LSTM initialize with word2vec \shortcite{dai2015semi}  & 10.00\%       \\
Full+Unlabeled+BoW \shortcite{maas-EtAl:2011:ACL-HLT2011} & 11.11\% \\
WRRBM+BoW (bnc) \shortcite{maas-EtAl:2011:ACL-HLT2011}  & 10.77\% \\
NBSVM-bi   \shortcite{wang2012baselines}           & 8.78\% \\
seq2-bow\emph{n}-CNN  \shortcite{johnson2014effective} & 7.67\% \\
Paragraph Vectors \shortcite{le2014distributed}             & 7.42\%        \\
LM-LSTM \shortcite{dai2015semi}     & 7.64\%        \\
SA-LSTM \shortcite{dai2015semi}     & 7.24\%        \\
\hline
SSVAE-I 				    & 9.37\% \\
SSVAE-II                                     & 8.72\%        \\
SSVAE-II,LM                               & 7.23\%        \\
\hline
\end{tabular}
\label{tab:imdb}
\end{table}

\subsection{Analysis of Conditional LSTM Structures}
From Table~\ref{tab:imdb_detail} and Table~\ref{tab:ag_detail}, the model with CLSTM-II outperforms CLSTM-I slightly.
We suppose that CLSTM-II receives label information more directly than CLSTM-I and hence can learn to differentiate various categories much easier.
Both of them surpass the model using vanilla LSTM evidently.

\begin{figure}[t]
    \centering
    \includegraphics[width=3in,height=2.2in]{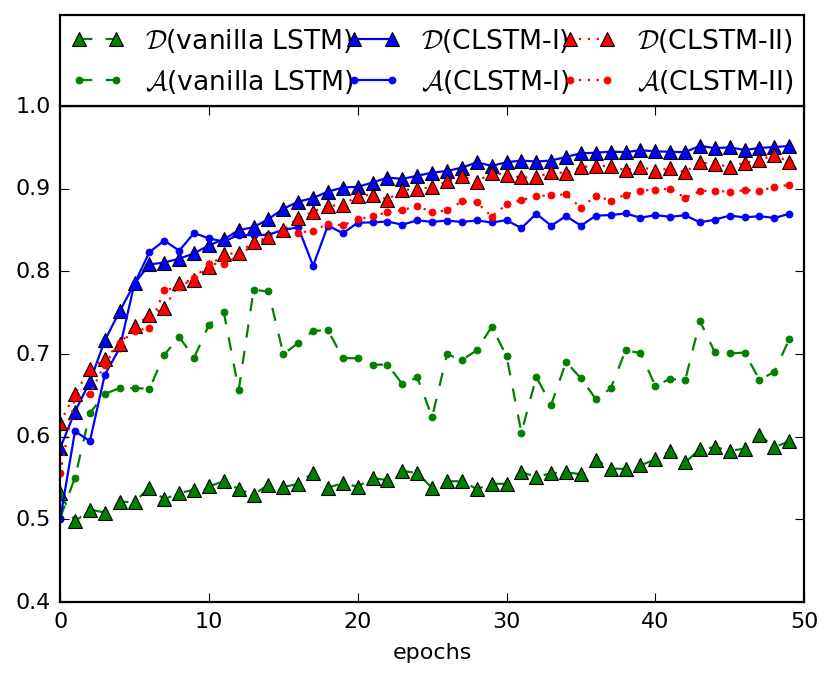}
    \caption{The discrimination index of decoder and classification accuracy between models using vanilla LSTM and conditional LSTMs, with 5K labeled data samples.}
    \label{fig:sclstm}
\end{figure}

To obtain a better understanding of these structures, we investigated the model using vanilla LSTM, CLSTM-I or CLSTM-II as its decoder quantitatively.
At first we define the following index for the decoder to explore its relationship with classification performance:
\begin{equation}
\mathcal{D} = \frac{1}{N_l}\sum_{i=1}^{N_l} 1\{\arg \max_y -\mathcal{L}(x^{(i)},y)=y^{(i)}\}\,,
\end{equation}
where $(x^{(i)},y^{(i)})$ is a sample in labeled set, $N_l$ is the number of total labeled data and $-\mathcal{L}(x, y)$ is the lower bound in Equ.~\ref{equ:lab}.
This equation denotes the ratio of samples that the decoder can produce higher evidence lower bound of generative likelihood with correct labels, in other words, ``how many rewards are given correctly''.
We use this index to evaluate decoder's discrimination ability.
The curves of models using these conditional LSTMs, together with classification accuracy $\mathcal{A}$, are shown in Fig.~\ref{fig:sclstm}.

By using CLSTMs, the accuracy improves rapidly as well as $\mathcal{D}$ index, which indicates the strong correlation between the accuracy of classifier and discrimination ability of conditional generative model.
At the early phase of training, the accuracy of vanilla LSTM improves quickly as well, but diverges at epoch 13.
Meanwhile the $\mathcal{D}$ index improves very slowly, indicating that not enough guiding information is provided by the decoder.
Therefore, the classifier is unable to utilize the unlabeled data to improve the accuracy, resulting in an unstable performance.

\subsection{Sampling with Baseline Methods}

Table~\ref{tab:imdb_detail} and ~\ref{tab:ag_detail} also list the results of models trained by sampling, as described in Sec.~\ref{sec:sample}.
In the implementation, sampling number $K$ is set to 1 when using S1 and 2 for S2.
The results of S2 for IMDB dataset are omitted, since the models using S2 are similar with the \emph{SSVAE}s on the datasets with only two classes.
The \emph{SSVAE-II} is used as the basic model for this comparison.

Experimental results demonstrate that the sampling-based optimization is made available by using two proposed baselines.
However, the models using S1 perform worse than the models without sampling, indicating that the variance is still high even using this baseline.
By using S2, the models achieve the performance on par with the \emph{SSVAE}s without sampling, which verifies the S2 as an efficient baseline method for \emph{SSVAE}s.
Besides, the time cost using S1 or S2 is less than that without sampling on both IMDB dataset and AG's News corpus (cf. Table~\ref{tab:time}).
For both S1 and S2, the adoption of pre-trained weights makes the optimization more stable during the experiments.

\begin{table}
\small
\caption{Time cost of training 1 epoch using different optimization methods on Nvidia GTX Titan-X GPU.}
\begin{tabular} {l|c|c|c}
\hline
Method & SSVAE-II & SSVAE-II,S1 & SSVAE-II,S2 \\
\hline
IMDB,2.5K & 4050(s) & 2900(s) & - \\
AG,8K & 1880(s) & 1070(s) & 1205(s) \\
\hline
\end{tabular}
\label{tab:time}
\end{table}

\begin{table*}
\center
\caption{Nice generated sentences conditioned on different categorical label $y$ and same latent state $z$.}
\resizebox{0.9\textwidth}{!}{

\begin{tabularx}{\textwidth}{ X|X }
\hline
\textbf{Negative} & \textbf{Positive} \\
\hline
this has to be one of the worst movies I've seen in a long time. &
this has to be one of the best movies I've seen in a long time. \\
\hline

what a waste of time ! ! !  &
what a great movie ! ! !  \\
\hline

all i can say is that this is one of the worst movies i have seen. &
anyone who wants to see this movie is a must see ! ! \\
\hline

UNK is one of the worst movies i've seen in a long time . &
UNK is one of my favorite movies of all time. \\
\hline

if you haven't seen this film , don't waste your time ! ! !  &
if you haven't seen this film , don't miss it ! ! ! \\

\hline

suffice to say that the movie is about a group of people who want to see this movie , but this is the only reason why this movie was made in the united states . &
suffice to say that this is one of those movies that will appeal to children and adults alike , but this is one of the best movies i have ever seen . \\
\hline

\end{tabularx}
}
\label{tab:gen_imdb}
\end{table*}

\subsection{Generating Sentences from Latent Space }

\begin{figure}
    \centering
    \includegraphics[width=3in,height=1.9in]{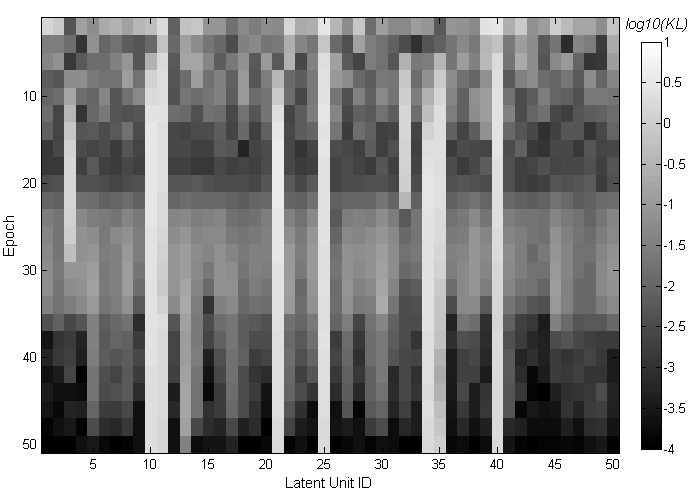}
    \caption{$\log D_{KL}(q_{\phi}(z|x,y) || p(z))$ for each latent unit is shown at different training epochs. High $KL$ (white) unit carries information about the input text $x$. }
    \label{fig:ztrain}
\end{figure}

To investigate whether the model has utilized the stochastic latent space, we calculated the $KL$-divergence for each latent variable unit $z_i$ during training, as seen in Fig.~\ref{fig:ztrain}.
This term is zero if the inference model is independent of the input data, i.e., $q_{\phi}(z_i|x,y)=p(z_i)$, and hence collapsed onto the prior carrying no information about the data.
At the end of training process, about 10 out of 50 latent units in our model keep an obviously non-zero value, which may indicate that the latent variable $z$ has propagated certain useful information to the generative model.

\begin{figure}[ht]
    \centering
    \includegraphics[width=3in,height=1.9in]{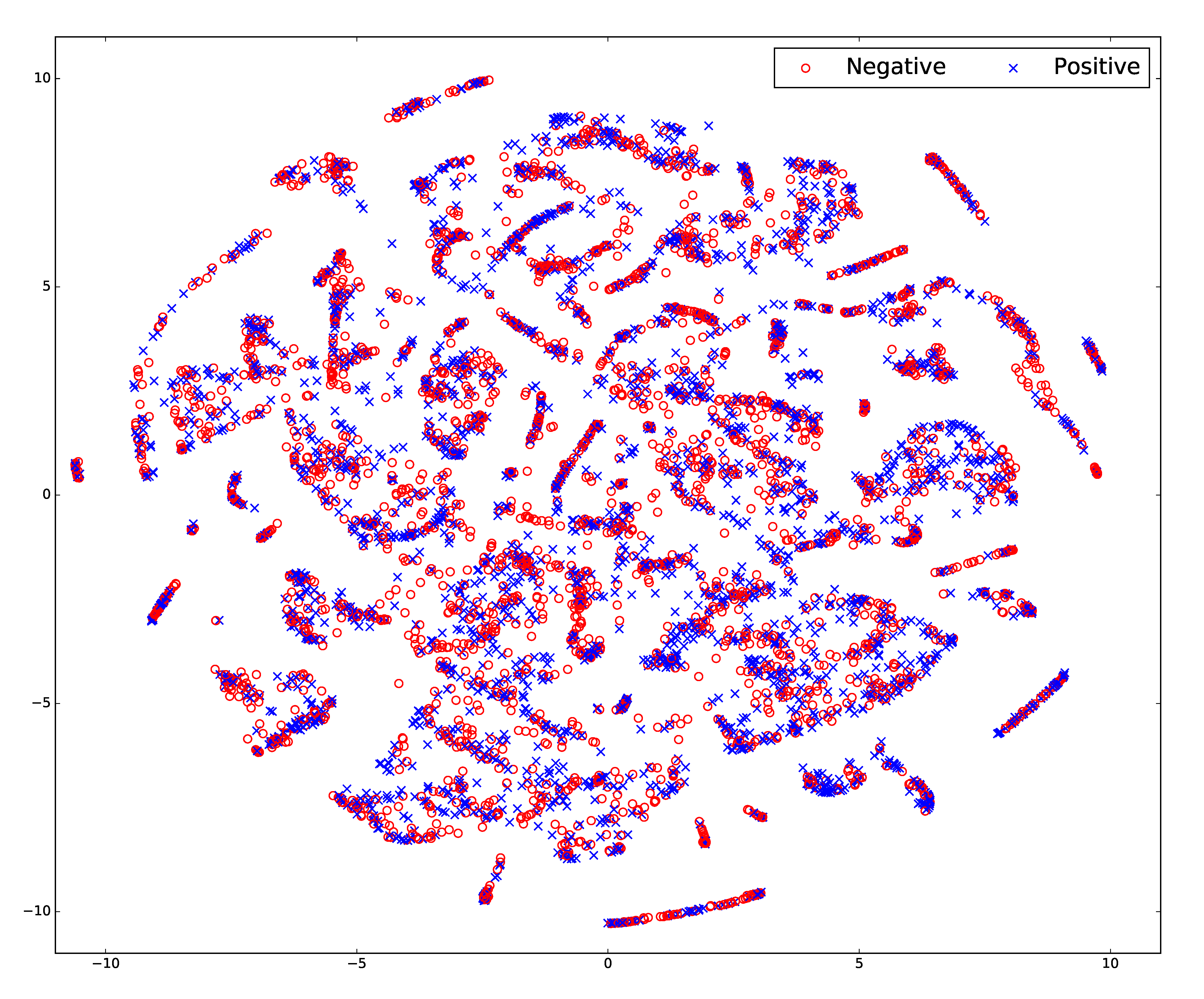}
    \caption{The distribution of IMDB data set in latent space using t-SNE. }
    \label{fig:tsne-imdb}
\end{figure}

To qualitatively study the latent representations, t-SNE~\cite{maaten2008visualizing} plots of $z \sim q_{\phi}(z|x,y)$ from IMDB dataset are seen in Fig.~\ref{fig:tsne-imdb}.
The distribution is Gaussian-like due to its normal prior $p(z)$ and the distributions of two classes are not separable.
When digging into some local areas (cf. supplementary materials), it's interesting to discover that sentences sharing similar syntactic and lexical structures are learned to cluster together, which indicates that the shallow semantic context and the categorical information are successfully disentangled. 

Another good explorative evaluation of the model's ability to comprehend the data manifold is to evaluate the generative model.
We selected several $z$ and generate sentences for IMDB using trained conditional generative model $p_\theta(x|y,z)$.
Table~\ref{tab:gen_imdb} demonstrates several cases using the same latent variable $z$ but with opposite sentimental labels.
Sentences generated by the same $z$ share a similar syntactic structure and words, but their sentimental implications are much different from each other.
The model seems to be able to recognize the frequent sentimental phrases and remember them according to categorical label $y$.
While faced with the difficulty for a model to understand real sentiment implication, it is interesting that some sentences can even express the sentimental information beyond the lexical phrases, e.g., ``\emph{but this is the only reason why this movie was made in the United States}''.
Similar interesting sentences can be also generated on AG's News dataset.

\section{Conclusion}
The \emph{SSVAE} has been proposed for semi-supervised text classification problem.
To explain why \emph{SSVAE} fails if using vanilla LSTM as its decoder, we provided an angle for \emph{SemiVAE}  from the perspective of reinforcement learning.
Based on this interpretation, the label information is enhanced in the \emph{SSVAE} by feeding labels to the decoder RNN  at each time step. This minor modification brings the \emph{SSVAE} into effect.
Two specific conditional LSTMs, i.e., CLSTM-I and CLSTM-II, are investigated.
Experimental results on IMDB dataset and AG's News corpus demonstrate that our method can achieve competitive performance compared with previous advanced models, and achieve state-of-the-art results by combining pretraining method.
In addition, the sampling-based optimization method has been proposed to reduce the computational complexity in training.
With the help of the baseline methods suggested in this paper, the model can be trained faster without loss of accuracy.

\section*{Acknowledgments}
This work was supported by National Key Basic Research Development Plan (973 Plan) Project of China under grant no. 2015CB352302, and partially supported by the Natural Science Foundation of China (NSFC) under grant no. 61375119 and no. 61673025, and Beijing Natural Science Foundation (4162029).

\bibliography{xuweidi}
\bibliographystyle{aaai}
\appendix

\end{document}